\documentclass{article}
\usepackage{ijcai20}

\usepackage{times}

\usepackage{booktabs}
\usepackage{url}  
\usepackage{graphicx}  

\usepackage{wrapfig}
\usepackage{multicol}
\usepackage{latexsym}
\usepackage{amsmath}
\usepackage{amssymb}
\usepackage{url}
\usepackage[noend]{algorithmic}
\usepackage{algorithm}
\usepackage{multirow}
\usepackage{mathtools}
\usepackage{enumitem}
\usepackage{color}
\usepackage{microtype}
\usepackage{tabularx}
\usepackage{amsthm}
\usepackage{bm}
\usepackage{amsthm}
\usepackage{epstopdf}
\usepackage[algo2e]{algorithm2e} 
\usepackage{subcaption}

\newtheoremstyle{mytheoremstyle} 
{\topsep}                    
{\topsep}                    
{\itshape}                   
{}                           
{\scshape}                   
{.}                          
{.5em}                       
{\thmname{#1}\thmnumber{ #2}\thmnote{ (#3)}}  

\theoremstyle{mytheoremstyle}

\newcommand{\Y}{\mathbb{Y}}
\newcommand{\Ycal}{\mathcal{Y}}

\newcommand{\irl}{IRL}

\newcommand{\Dcal}{\mathcal{D}}

\title{Maximum Entropy Multi-Task Inverse RL
}
\author{
	Saurabh Arora$^1$\footnote{Contact Author}\and
	Bikramjit Banerjee$^2$\and
	Prashant Doshi$^1$\\
	\affiliations
	$^1$
	{THINC Lab, Dept. of Computer Science,}
	{415 Boyd GSRC,}
	{University of Georgia, Athens,} 
	{GA-} 
	{30602}
	\\
	$^2$
	{School of Computing Sciences \& Computer Engineering,}
	{118 College Drive,}
	{University of Southern Mississippi, Hattiesburg,} 
	{MS-} 
	{39406}
	\\
	\emails
	sa08751@uga.edu,
	Bikramjit.Banerjee@usm.edu,
	pdoshi@cs.uga.edu
}

\begin{document}

\maketitle

\begin{abstract}

  Multi-task \irl{} allows  for the possibility that  the expert could
  be switching between  multiple ways of solving the  same problem, or
  interleaving demonstrations  of multiple tasks. The  learner aims to
  learn the multiple reward functions that guide these ways of solving
  the  problem. We  present a  new method  for multi-task  \irl{} that
  generalizes  the  well-known  maximum  entropy approach  to  IRL  by
  combining  it with  the Dirichlet  process based  clustering of  the
  observed input. This yields a single nonlinear optimization problem,
  called  MaxEnt Multi-task  \irl{},  which can  be  solved using  the
  Lagrangian  relaxation and  gradient  descent  methods. We  evaluate
  MaxEnt  Multi-task  \irl{} in  simulation  on  the robotic  task  of
  sorting  onions  on a  processing  line  where the  expert  utilizes
  multiple ways of detecting and removing blemished onions. The method
  is able to learn the underlying  reward functions to a high level of
  accuracy and it improves on the previous approaches to multi-task
  \irl{}.
\end{abstract}


\section{Introduction}
\label{sec:intro}
Inverse  reinforcement  learning
(IRL)~\cite{Ng00:Algorithms,Russell98:Learning} refers  to the problem
of  ascertaining  an  agent's  preferences from  observations  of  its
behavior  while executing  a task.
For instance, observing  a  human  perform  a  task on  the  factory  line  provides
information and  facilitates learning the  task. This passive  mode of
transferring  skills  to a  collaborative  robot  (cobot) is  strongly
appealing because  it significantly  mitigates costly human  effort in
not only manually programming the  cobot but also in actively teaching
the  cobot  through  interventions.   
The learned preferences can
be    utilized    by   the    cobot    to    imitate   the    observed
task~\cite{Osa18:Algorithmic}     or    assist     the    human     on
it~\cite{Trivedi18:Inverse}.

Motivated  by the  long-term goal  of bringing  robotic automation  to
existing produce  processing lines, we  focus on the  well-defined but
challenging task of sorting onions. 
Our observations of persons engaged  in this job  in a real-world produce  processing shed
attached  to a  farm revealed  multiple sorting  techniques in  common
use. For  example, in addition to  the overt technique of  picking and
inspecting few  of the  onions as  they pass by,  we noticed  that the
humans  would simply  roll the  onions  (without picking  them up)  to
expose more of their surface.  The latter technique allows more onions
to be quickly assessed but inaccurately.
Consequently,  the  problem  of  learning  how  to  sort  onions  from
observations  requires   multi-task  IRL~\cite{Arora18:Survey}.   This
variant of IRL  allows the possibility that the  demonstrator could be
switching  between   multiple  reward  functions   thereby  exhibiting
multiple  ways of  solving the  given problem  or performing  multiple
tasks. In a previous approach to  multi-task Bayesian IRL, DPM-BIRL  \cite{Choi12:Nonparametric}, a  Dirichlet
process  model is  used to  perform non-parametric  clustering of  the trajectories, where  each cluster corresponds to  an underlying reward function.     
Different   from    Bayesian   IRL,    Babes-VRoman   et al. ~\cite{Babes-Vroman11:Apprenticeship} apply  the iterative EM-based clustering by  replacing the  mixture of Gaussians  with a  mixture of reward  functions  and a reward  function  of maximum  likelihood  is learned for each cluster. 

Unlike previous approaches,  we  present  a new  method  for  multi-task IRL  that
generalizes   the  well-known   maximum   entropy   approach  to   IRL
(MaxEntIRL)~\cite{Ziebart08:Maximum}.             
Arora            and Doshi~[\citeyear{Arora18:Survey}] list  MaxEntIRL  as  a  key  foundational
technique  for IRL  in  their survey. Indeed,  applications of  this
technique    in    various    contexts    have    yielded    promising
results~\cite{Bogert14:Multi,Wulfmeier2015}.      A    straightforward
extension  of MaxEntIRL  to multiple  tasks  would be  to replace  the
maximum likelihood computation in the  iterative EM with the nonlinear
program of  MaxEntIRL.  But this  would require solving  the nonlinear
program for each  of multiple clusters, and  repeatedly.  In contrast,
we formulate the  problem as a single  entropy-based nonlinear program
that combines the MaxEntIRL objective  with the objective of finding a
cluster assignment distribution having the least entropy. Ziebart et al.~[\citeyear{Ziebart08:Maximum}] demonstrated the advantage that MaxEntIRL brings to single-task IRL in comparison to the Bayesian technique. We expect to leverage this benefit toward multi-task IRL. 


Modeling multi-task IRL  as a single optimization  problem enables the
direct  application of  well-studied optimization  algorithms (e.g. fast gradient-descent) to  this problem. In  particular, we  derive the  gradients of  the Lagrangian relaxation of the nonlinear program, which then facilitates the use of fast gradient-descent  based algorithms.  We evaluate  the performance of this  method in comparison with two previous multi-task IRL techniques  on a simulation  in ROS  Gazebo of  the onion  sorting problem. We show that the MaxEnt multi-task method improves on both and learns the reward functions to  a high level
of accuracy,  and which  allows Sawyer to  observe and  reproduce both
ways of  sorting the  onions while making  few mistakes.   However, we
also observed room for improvement in one of the learned behaviors.

\section{Background}
\label{sec:background}

In \irl{}, the task of a learner is to find  a  reward  function under  which  the  observed behavior of expert, with dynamics modeled as an incomplete MDP $\langle S,A,T  \rangle$,  is  optimal ~\cite{Russell98:Learning,Ng00:Algorithms}. 
Abbeel and Ng~[\citeyear{Abbeel04:Apprenticeship}]
first suggested modeling  the reward function as  a linear combination
of  $K$   binary  features,  $\phi_k$:  $S   \times  A$  $\rightarrow$
$\{0,1\}$, $k \in \{1,2 \ldots K\} $,  each of which maps a state from
the set of states  $S$ and an action from the  set of expert's actions
$A$ to a  value in \{0,1\}.  The reward function is then
defined                                                             as
$R(s,a)  =   \bm{\theta}^T  \phi(s,a)  =   \sum_{k=1}^K  \theta_k\cdot
\phi_k(s,a)$, where $\theta_k$ are the {\em feature weights} in vector
$\bm{\theta}$. The  learner's task is to
find  a vector  $\bm{\theta}$  that  completes the  reward function,  and
subsequently, the MDP such that the observed behavior is optimal.

Many of the early methods for IRL biased their search for the solution
to combat the ill-posed nature of IRL and the very large search space \cite{Abbeel04:Apprenticeship,Ziebart08:Maximum}. Ziebart   et   al.~[\citeyear{Ziebart08:Maximum}], taking a contrasting perspective,  
sought to find a  distribution over all trajectories (sequences of state-action pairs) that
exhibited the  maximum entropy  while being  constrained to  match the
observed  feature  counts.  The problem reduces  to finding 
$\bm{\theta}$, which  parameterizes the exponential  distribution that     exhibits    the     highest     likelihood. 
The  corresponding  nonlinear  program is  shown
below.
\begin{align}
\begin{array}{l}
  \max \nolimits_{\Delta} ~~ -\sum\nolimits_{i=1}^{|\Y|} P(y_i)~log
  P(y_i)\\
  \mbox{{\bf subject to}}~~~\\
  \sum \nolimits_{i=1}^{|\Y|} P(y_i) = 1 \\
  E_{\Y}[\phi_k]  = \hat{\phi}_k ~~~~~~ \forall k 
\label{eq:ziebart-max-ent}
\end{array}
\end{align}
Here, $\Delta$ is the space of  all distributions over the set $\Y$ of
all                          trajectories,                         and
$E_{\Y}[\phi_k]=\sum  \nolimits_{i=1}^{|\Y|} P(y_i)~\sum_{(s,a)  \in  y_i}
\phi_k(s,a)$.    Let   $\Ycal$    denote   the    set   of    observed
trajectories. Then, the right-hand side of the second constraint above
becomes,
$\hat{\phi}_k = \frac{1}{|\Ycal|}\sum_{i=1}^{|\Ycal|} \sum_{(s,a) \in y_i}
\phi_k(s,a)$.

\subsection{Multi-Task \irl{}}
\label{subsec:dpm_multitask}

The same job on a processing line may be performed in one of many ways
each guided by  a distinct set of  preferences. An expert may
switch  between these  varied behaviors  as  it performs  the job,  or
multiple experts  may interleave  to perform the  job.  If  the reward
functions  producing  these behaviors  are  distinct,  then using  the
traditional  IRL methods  would yield  a single  reward function  that
cannot  explain the  observed trajectories  accurately.  However,  modeling it as multi-task  problem  allows  the
possibility of learning multiple reward functions.
If the number  of involved reward functions is  pre-determined, we may
view each  unknown reward function  as a generative model  producing a
cluster of trajectories  among the observed set.  As  both, the reward
weights and set of observed trajectories 
generated  by   it,  are  unknown,   we  may  utilize   the  iterative
EM                    to                   learn
both~\cite{Babes-Vroman11:Apprenticeship}.  

On the  other hand,  if the  number of reward  functions is  not known
a'priori, multi-task  IRL can be viewed  as non-parametric
mixture model clustering,  which is typically anchored  by a Dirichlet
process~\cite{Gelman13:Bayesian}.  We  briefly  review  the  Dirichlet
process      and  outline   its   previous   use  in DPM-BIRL~\cite{Choi2013bayesian}.

A Dirichlet  Process (DP) is  a stochastic process whose  sample paths
are  drawn  from  functions  that are  distributed  according  to  the
Dirichlet distribution.  More formally, function $G \sim DP(\alpha,H)$
if    $(G(X_1),   G(X_2),    \ldots,    G(X_r))$   $\sim$    Dirichlet
$(\alpha   H(X_1),   \alpha    H(X_2),\ldots,\alpha   H(X_r))$   where
$\{X_1, X_2, \ldots, X_r\}$ is some  partition of the space over which
base distribution $H$ is defined.  Here, $\alpha$ is the concentration
parameter of the Dirichlet distribution  and 
$E[G(X_r)]    =
H(X_r)$. Observations distributed according to  $G$ allow us to update
the DP.  Let  $\theta_1, \theta_2, \ldots, \theta_D$ be  a sequence of
observations and $n_r$ be the number  of these observations that lie in
$X_r$.          Then,        the         posterior        distribution
$(G(X_1),             G(X_2),            \ldots,             G(X_r))|$
$\theta_1,       \theta_2,        \ldots,       \theta_D$       $\sim$
Dirichlet$(\alpha  H(X_1)+n_1,   \alpha  H(X_2)+n_2,   \ldots,  \alpha
H(X_r)+n_r)$.

DPs     find     application     in     Bayesian     mixture     model
clustering~\cite{Gelman13:Bayesian}  due  to an  interesting  property
exhibited by $G$.   Irrespective of whether the  base distribution $H$
is smooth, $G$  is a discrete distribution.   Therefore, i.i.d.  draws
of  $\theta_D$ from  $G$ as  $D \rightarrow  \infty$ will  repeat, and
these  can  be seen  as  cluster  assignments. 
A generic DP-based Bayesian mixture model can be defined as:
{\small$$G|\alpha,H \sim DP(\alpha,H);\quad \theta_i|G \sim G;\quad y_i|\theta_i \sim F(\theta_i)$$}
where data  $y_i$ has distribution  $F(\theta_i)$. Notice the  lack of
any bound on the number of mixture components. To utilize this mixture
model  for clustering  observed data  $\{y_i\}$, we  must additionally
assign  each  data  point  to   its  originating  cluster,  and  these
assignments     are    drawn     from    convex     mixture    weights
($\pi_1,  \pi_2,  \ldots,  \pi_D$) which  are  themselves  distributed
randomly. The  number of components $D$  may grow as large  as needed.
The  popular  stick-breaking  construction  imposes  the  distribution
$\beta_d\prod_{l=1}^{d-1} (1  - \beta_l)$ to a  mixture weight $\pi_d$
where parameter $\beta_d$  is sampled from the  Beta distribution.  We
may then obtain  a cluster assignment $c_i$ for data  point $y_i$ by
sampling   distribution $G$  parameterized  by   event
probabilities $\bm{\pi}$. 
{\small\begin{align*}
&\bm{\beta} \sim Dirichlet(\frac{\alpha}{D}, \ldots,
\frac{\alpha}{D}); & \pi_d = \beta_d \prod_{l=1}^{d-1} (1 - \beta_l) ~~~ \forall d.
\end{align*}}
Then, for each data point
{\small\begin{align}
&\theta^*_d |H \sim H; 
&c_i|\bm{\pi} \sim G;\quad
&y_i|c_i, \{\theta^*_d\} \sim F(\theta^*_{c_i})
\label{eqn:DP_clustering}
\end{align}}
where  $\theta^*_d$ denotes  a unique  component parameter  value. 

Choi  and Kim~[\citeyear{Choi2013bayesian}]  utilize  this Bayesian  mixture
model  application of  a DP  toward multi-task  IRL.  The  data points
$\{y_i\}$ are the observed trajectories, $\{\theta^*_d\}$ parameterize
the $D$ distinct reward functions, and $F(\theta^*_{c_i})$ corresponds
to                                       $\frac{1}{Z(\theta^*_{c_i})}$
$e^{\tiny\sum\nolimits_{t=1}^T  Q(s_t,a_t;\theta^*_{c_i})}$,  where $T$  is
the  fixed length  of the  trajectory, $Z(\theta^*_{c_i})$ is  
partition  function, and $H$   is   taken    as   the   Gaussian   distribution.
Neal~[\citeyear{Neal00:Markov}]  discusses   several  MCMC   algorithms  for
posterior  inference on  DP-based  mixture models,  and  Choi and  Kim
selected the Metropolis-Hastings. 

\section{MaxEnt Multi-Task \irl{}}
\label{sec:maxent_multitask}

Ziebart  et al.~[\citeyear{Ziebart08:Maximum}]  notes a  key benefit  of the
MaxEnt     distribution    over     trajectories    
over  the distribution $F(\theta^*_{c_i})$
mentioned  in Section~\ref{subsec:dpm_multitask}  (which is  the prior
over   trajectories  utilized   in   the   Bayesian  formulation   for
\irl{}~\cite{Ramachandran07:Bayesian}),    despite    their    initial
similarities.  Specifically, the  latter formulation, which decomposes
the trajectory into its constituent state-action pairs and obtains the
probability of each state-action  as proportional to the exponentiated
Q-function, is  vulnerable to the label  bias. Due to the  locality of
the action probability computation, the distribution over trajectories
is impacted by  the number of  action choice  points (branching)
encountered  by  a   trajectory.   On  the  other   hand,  the  MaxEnt
distribution does not suffer from this bias.

This  important  observation motivates  a  new  method that combines the non-parametric clustering of trajectories and
learning   multiple   reward   functions   by  finding   trajectory
distributions of maximum entropy.  This  method would have the benefit
of avoiding  the label bias,  which afflicts  the previous
technique of DPM-BIRL.

\subsection{Unified Optimization}
\vspace{-0.01in}

A straightforward approach to  the combination would be  to  replace the parametric distribution  in MaxEnt\irl{}  with $F(\theta^*_{c_i})$ in DP-based mixture model,  the  distribution  over  the   trajectories  
with cluster assignment value $c_i$.  Solving the nonlinear program will yield parameter $\theta^*_{c_i}$ that maximizes  the entropy of
$F(\theta^*_{c_i})$. 
Though simple,  this approach is inefficient because  it requires
solving  the  MaxEnt program  repeatedly  --  each time  the  DP-based
mixture model is  updated. As an analytical solution of  MaxEnt is not
available,  the optimization  is  performed  numerically by using either  
gradient  descent ~\cite{Ziebart08:Maximum} or  
 L-BFGS ~\cite{Bogert14:Multi}. 

Instead, we pursue an approach that  adds key elements of the DP-based
mixture   modeling  to   the   nonlinear  program   of  MaxEnt optimization. MaxEnt can learn component
parameters  $\{\theta_d\}$  (these  are the  Lagrangian  multipliers),
which  maximize the  entropy  of the  distribution $F(\theta_d)$  over
those trajectories whose cluster  assignment $c_i = d$.  Subsequently,
each   component   distribution   $F$   assumes   the   form   of   an
exponential-family distribution parameterized  by $\theta_d$, which is
known  to exhibit  the  maximum  entropy. For our DPM model, the distribution $G$ is mixture $ \tiny\sum\nolimits_{d =1}^D \pi_d \delta_{\theta^*_d}$. To  multi-task max-entropy objective,  we  add a  second
objective  of  finding component  weights  $\bm{\pi}$  that exhibit  a
minimal entropy.  The effect of this objective is to learn a minimal number of  distinct clusters. More formally, the objective function is

\begin{align*}
&\max \nolimits_{P(y_i|c_i) \in \Delta, \bm{\pi} \in \Delta} 
-  \sum\nolimits_{d=1}^D   \sum\nolimits_{i=1}^{|\Y|}
P(y_i|c_i=d) \\
&\log P(y_i|c_i=d)
+   \sum\nolimits_{d=1}^D  \pi_d~\log\pi_d.
\end{align*}

\noindent $P(y_i|c_i=d)$ here can be written as $\delta_d(c_i) Pr_d(y_i)$ where $\delta_d(c_i)$ is the Kronecker delta  taking a value of 1 when
$c_i = d$,  and 0 otherwise, and $Pr_d(y_i)$ is  the distribution over
all  the   trajectories  for  cluster  $d$.    The  unified  nonlinear
optimization problem is shown below. 
\begin{align}
&\max \nolimits_{Pr_d(y_i) \in \Delta^D, \bm{\pi} \in \Delta} 
- \sum\nolimits_{d} \sum\nolimits_{i=1}^{|\Y|}
\delta_d(c_i) Pr_d(y_i)~\log \delta_d(c_i)\nonumber\\
&  Pr_d(y_i)+  \sum\nolimits_{d} \pi_d~\log\pi_d\nonumber\\
&\mbox{{\bf subject to}}~~~\nonumber\\
&\tiny\sum\nolimits_{d} \tiny\sum\nolimits_{y_i \in \Y} P(y_i,c_i=d) = 1\nonumber\\ 
&E_{\Y}[\phi_k|c_i=d]  = \hat{\phi}_{d,k} ~~~~~~ \forall d \in D,
\forall k \in K \nonumber\\
&\sum \nolimits_{d=1}^\Dcal{} \pi_d = 1
\label{eq:max-ent-min-ent}
\end{align}

The first constraint  above simply ensures that  the joint probability
distribution  sums to  1. The  second constraint  makes the  analogous
constraint  in~MaxEnt\irl{}  more specific  to  matching
expectations of feature  functions that belong to  the reward function
of cluster $d$.  Here, {\small\begin{align*}
&E_{\Y}[\phi_k|c_i=d] = \tiny\sum       \nolimits_{i=1}^{|\Y|}
P(y_i,c_i=d)~\tiny\sum\nolimits_{(s,a) \in y_i} \phi_k(s,a)\nonumber\\
& =\tiny\sum       \nolimits_{i=1}^{|\Y|}
P(y_i|c_i=d)~P(c_i=d)~\tiny\sum\nolimits_{(s,a) \in y_i} \phi_k(s,a)\nonumber\\
& =\pi_d\tiny\sum\nolimits_{i=1}^{|\Y|}
\delta_d(c_i) Pr_d(y_i)~\tiny\sum\nolimits_{(s,a) \in y_i} \phi_k(s,a),
\end{align*}}\vspace{-0.1in}
and \vspace{-0.1in}
{\small\begin{align*}
&\hat{\phi}_{d,k} =
\frac{1}{|\Ycal|}\tiny\sum\nolimits_{i=1}^{|\Ycal|}\delta_d(c_i)
\tiny\sum\nolimits_{(s,a) \in y_i} \phi_k(s,a).  
\end{align*}}
Constraint 3 of the  program in~\eqref{eq:max-ent-min-ent} ensures that
the mixture weights are convex. Notice that the DP-based mixture model
  obtains    cluster  assignment
$c_i$ from mixture  weights $\bm{\pi}$.  We  may approximate  this
simulation           of            $c_i$           simply           as
$\pi_d = \frac{1}{|\Ycal|} \tiny\sum\nolimits_{i=1}^{|\Ycal|} \delta_d(c_i)$,
which is the proportion of observed trajectories currently assigned to
cluster   $c_i$.    For   notational  convenience,   let   us   denote
$\delta_d(c_i)$  as indicator $v_{d,i}$. 
We may  then rewrite  the first constraint as 
{\small\begin{align*}
& \sum\nolimits_{d}  \sum\nolimits_{y_i=1}^{|\Y|}
P(y_i|c_i=d)\pi_d = 1\\
 \Leftrightarrow &\sum\nolimits_{d} \sum\nolimits_{y_i=1}^{|\Y|}
  P(y_i|c_i=d)\pi_d = 1\\
\Leftrightarrow & \sum\nolimits_{d} \pi_d  \sum\nolimits_{y=1}^{|\Y|}
P(y_i|c_i=d) = 1\\
\Leftrightarrow & \sum\nolimits_{d} \pi_d  \sum\nolimits_{i=1}^{|\Y|}
\delta_d(c_i) Pr_d(y_i) = 1\\
\Leftrightarrow &  \sum\nolimits_{d}
\frac{1}{|\Ycal|} \sum\nolimits_{i=1}^{|\Ycal|}
v_{d,i}   \sum\nolimits_{i=1}^{|\Y|} v_{d,i} Pr_d(y_i) = 1
\end{align*}}
and the second constraint is rewritten as,
{\small\begin{align}
E_{\Y,d}[\phi_{k}] = \frac{\sum\nolimits_{i=1}^{|\Ycal|}
v_{d,i}}{|\Ycal|} \sum\limits_{i=1}^{|\Y|}
v_{d,i} Pr_d(y_i)\sum\limits_{(s,a) \in y_i} \phi_{k}(s,a)
\label{eq:expectation}
\end{align} }
while
{\small\begin{align}
\hat{\phi}_{d,k}    =    \frac{1}{|\Ycal|}\sum\nolimits_{i=1}^{|\Ycal|}
v_{d,i} \sum\nolimits_{(s,a) \in y_i} \phi_{k}(s,a).
\label{eq:emp_expectation}
\end{align}}

Furthermore, we may simplify the third constraint of the nonlinear
program as follows:
{\small\begin{align*}
 \sum \nolimits_{d=1}^\Dcal{} \pi_d = 1 
\Leftrightarrow & \sum \nolimits_{d=1}^\Dcal{} \frac{1}{|\Ycal|}
 \sum\nolimits_{i=1}^{|\Ycal|} v_{d,i} = 1\\
\Leftrightarrow & \sum\nolimits_{i=1}^{|\Ycal|}  \sum \nolimits_{d=1}^\Dcal{} 
v_{d,i} = |\Ycal|\\ 
\Leftrightarrow & \sum \nolimits_{d=1}^\Dcal{} 
v_{d,i} = 1
\end{align*}}
The last equivalence follows from  the fact that every observed
trajectory must belong to a cluster, and $v_{d,i} \in \{0,1\}$. The final form of the NLP of~\eqref{eq:max-ent-min-ent} is as follows.
\begin{align}
&\max \nolimits_{Pr_d(y_i) \in \Delta^D, \bm{v}_{d,i} \in \{0,1\}} 
- \sum\nolimits_{d} \sum\nolimits_{i=1}^{|\Y|}
v_{d,i} Pr_d(y_i)~\log v_{d,i} \nonumber\\
& Pr_d(y_i)+  \frac{1}{|\Ycal|}\sum\nolimits_{d} \sum\nolimits_{i=1}^{|\Ycal|} v_{d,i}~\log\frac{1}{|\Ycal|}\sum\nolimits_{i=1}^{|\Ycal|} v_{d,i}\nonumber\\
&\mbox{{\bf subject to}}~~~\nonumber\\
&  \sum\nolimits_{d}
\frac{1}{|\Ycal|} \sum\nolimits_{i=1}^{|\Ycal|}
v_{d,i}   \sum\nolimits_{i=1}^{|\Y|} v_{d,i} Pr_d(y_i) = 1\nonumber\\ 
&E_{\Y,d}[\phi_{k}]  = \hat{\phi}_{d,k} ~~~~~~ \forall d \in D,
\forall k \in K \nonumber\\
& \sum\nolimits_{d}  v_{d,i} = 1,\forall {i\in \{1,\ldots,|\Y|\}}
\label{eq:max-ent-min-ent2}
\end{align}
where $E_{\Y,d}[\phi_{k}],\hat{\phi}_{d,k}$ are defined in Eqs.~\ref{eq:expectation}
and~\ref{eq:emp_expectation}.

\subsection{Gradient Descent}

The     Lagrangian     dual      for     the     nonlinear     program
in~\eqref{eq:max-ent-min-ent2}         is         optimized         as
$\max                  \nolimits_{Pr_d,                  \bm{v}_{d,i}}
\min\nolimits_{\eta,\theta_{d,k},\lambda_i} \mathcal{L}$ with

{\small
	\begin{align}
	\mathcal{L}&= \left(- \sum\nolimits_{d} \sum\nolimits_{i=1}^{|\Y|}
	v_{d,i} Pr_d(y_i)~\log v_{d,i} Pr_d(y_i) \right)\nonumber\\ 
	&+\left( \frac{1}{|\Ycal|}\sum\nolimits_{d} \sum\nolimits_{i=1}^{|\Ycal|} v_{d,i}~\log\frac{1}{|\Ycal|}\sum\nolimits_{i=1}^{|\Ycal|} v_{d,i} \right ) \nonumber\\
	&+\eta \left( \sum\nolimits_{d}
	\left(\frac{1}{|\Ycal|}\sum\nolimits_{i=1}^{|\Ycal|}
	v_{d,i} \right) \sum\nolimits_{i=1}^{|\Y|} v_{d,i} Pr_d(y_i) - 1 \right)\nonumber\\
	& + \sum\nolimits_{d}\sum\nolimits_{k} \bm{\theta}_{d,k} \bigg( \left(\frac{1}{|\Ycal|}\sum\nolimits_{i=1}^{|\Ycal|}
	v_{d,i} \right)  \sum \nolimits_{i=1}^{|\Y|}
	v_{d,i} Pr_d(y_i)\nonumber\\
	&~\sum\nolimits_{(s,a) \in y_i} \phi_{k}(s,a) - \frac{1}{|\Ycal|}\sum\nolimits_{i=1}^{|\Ycal|} v_{d,i} 
	\sum\nolimits_{(s,a) \in y_i} \phi_{k}(s,a) \bigg) \nonumber\\
	&+ \sum\nolimits_{i=1}^{|\Ycal|} \lambda_i \left( \sum\nolimits_{d}v_{d,i} -1\right)
	\end{align}
}%
where multipliers  $\eta,\{\lambda_i\}_{y_i\in\Y}$ can  be substituted
by using  relations derived  by equating derivatives  of $\mathcal{L}$
w.r.t.  variables  of optimization to 0.   The target is to  learn the
multipliers  $\{\bm{\theta}_d\}$ (weights  for linear  reward function
for each  learned cluster $d$)  and the variables $v_{d,i}$  (for each
trajectory        $y_i       \in        \Y$)       that        achieve
$\max   \nolimits_{Pr_d,  \bm{v}_{d,i}}   \min\nolimits_{\theta_{d,k}}
\mathcal{L}$. We achieve the target  via gradient ascent for $v_{d,i}$
and   descent  for   $\theta_{d,k}$   using   the  following partial
derivatives:
{\small
\begin{flalign*}
&\frac{\partial \mathcal{L}}{\partial \theta_{d,k}}
=E_{\Y,d}[\phi_k]-\hat{\phi}_{d,k}\\
\small
&\frac{\partial \mathcal{L}}{\partial v_{d,i}}
=\frac{\big(\tiny\sum    \nolimits_{i=1}^{|\Y|}
P(y_i|c_i=d)+1+\frac{|\Ycal|}{v_{d,i}}(1-    \log
Z(\theta_{d,k})\big)}{\sum\nolimits_{i=1}^{|\Ycal|}  v_{d,i}}
\end{flalign*}}
where      $P(y_i|c_i=d)=\frac{\exp     \left(\pi_d\sum\nolimits_{k=1}^K
	\theta_{d,k} \sum\nolimits_{(s,a)  \in y_i} \phi_{k}(s,a)\right)}{
	Z(\theta_{d,k})}$    and    $Z(\theta_{d,k})=\sum\nolimits_{d    \in
	\Dcal{}}\pi_d\sum      \nolimits_{i=1}^{|\Y|}\exp      $      $\big(
\pi_d\sum\nolimits_{k=1}^K$ $\theta_{d,k}  \sum\nolimits_{(s,a) \in y_i}
\phi_{k}(s,a)\big)$. The  former derivative is  same as that  used for
the  single  task  MaxEnt\irl{}.
The latter derivative indicates  that the chances of  change in assignment
is less if  a cluster has many trajectories assigned  to it (inversely
proportional             to             $\sum\nolimits_{i=1}^{|\Ycal|}
v_{d,i}$ ) and has a higher likelihood of generating trajectories. Due
to lack of space, we do not show the derivations of these gradients in
this paper.

\section{Domain: Robotic Sorting of Onions}
\label{sec:domain}

Our broader vision  is to make it easy to  deploy robotic manipulators
on complex  processing lines involving pick-inspect-place  tasks using
 \irl{}.  With this vision, we  seek to
deploy the  robotic arm  Sawyer for  sorting vegetables  in processing
sheds. Our  setup involves  a learner robot  observing an  expert sort
onions  in a  post-harvest processing  facility.  The  expert aims  to
identify  and remove  onions  with blemishes  from  the collection  of
onions present on  a conveyor belt. Blemished onions are  dropped in a
bin while others are allowed to  pass. We simulate
a human expert with another robot.
\begin{figure}
\centerline{
\includegraphics[width=0.116\textwidth,
			height=0.116\textwidth]{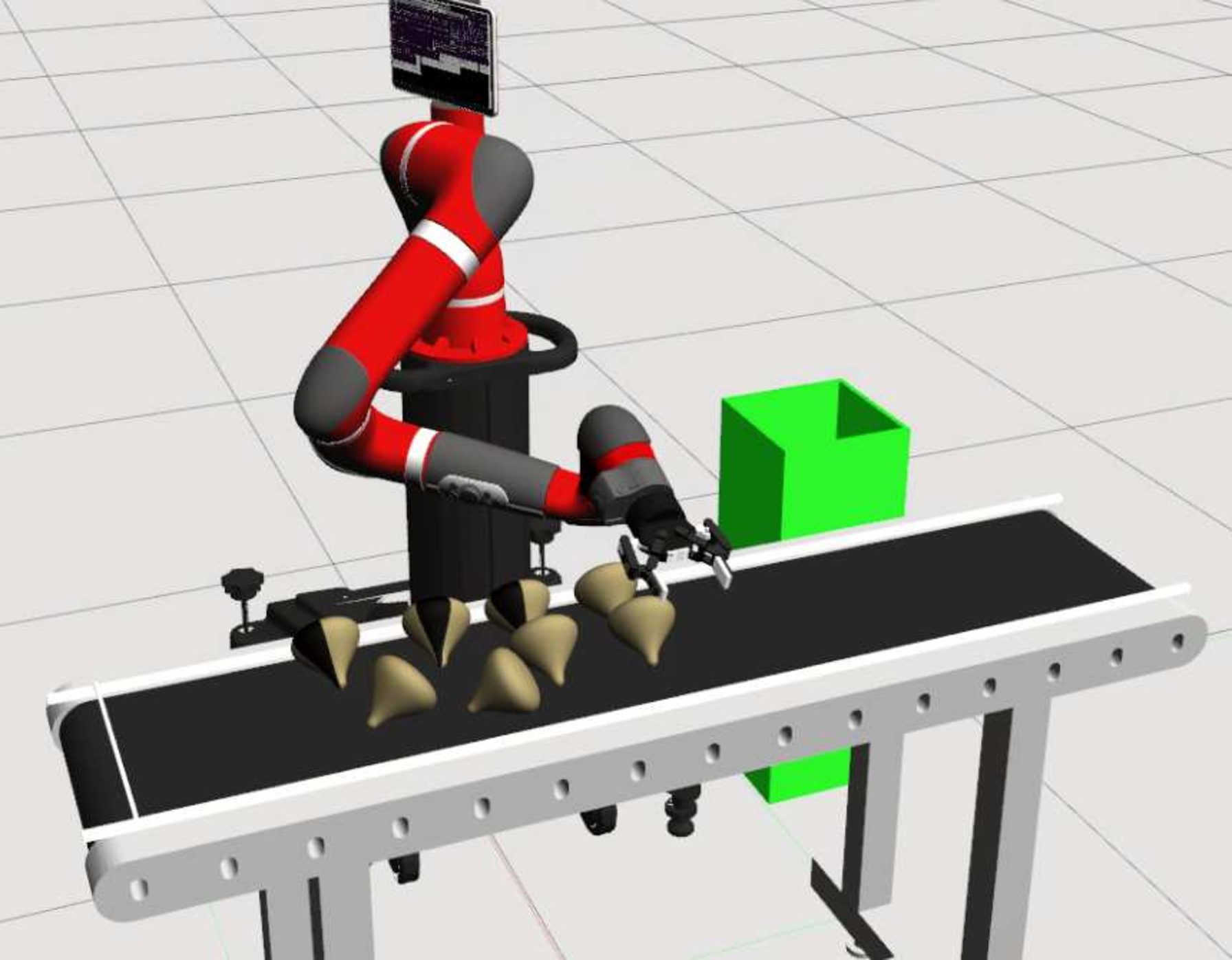}
			\includegraphics[width=0.116\textwidth,
			height=0.116\textwidth]{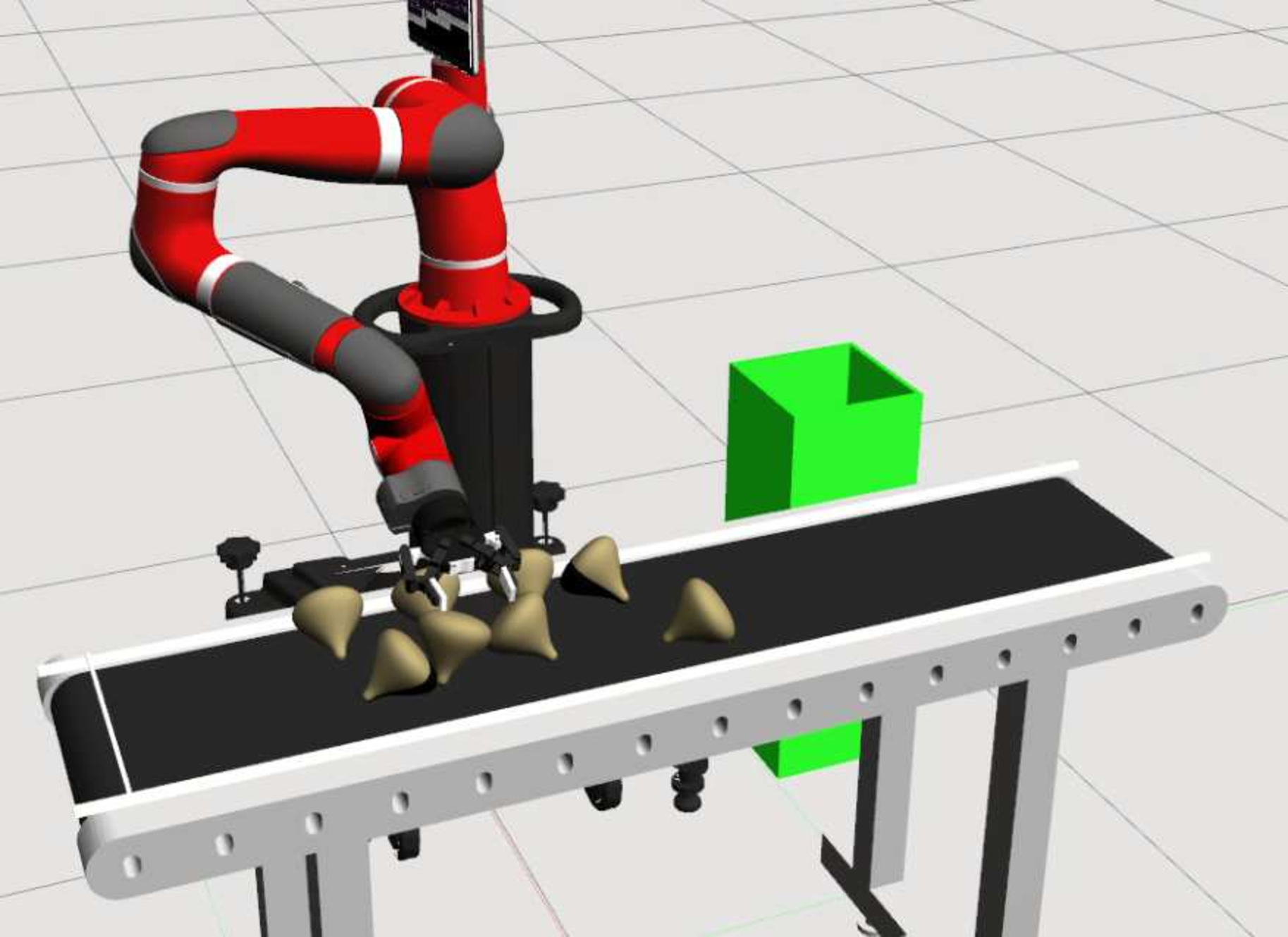}
			\includegraphics[width=0.116\textwidth,
			height=0.116\textwidth]{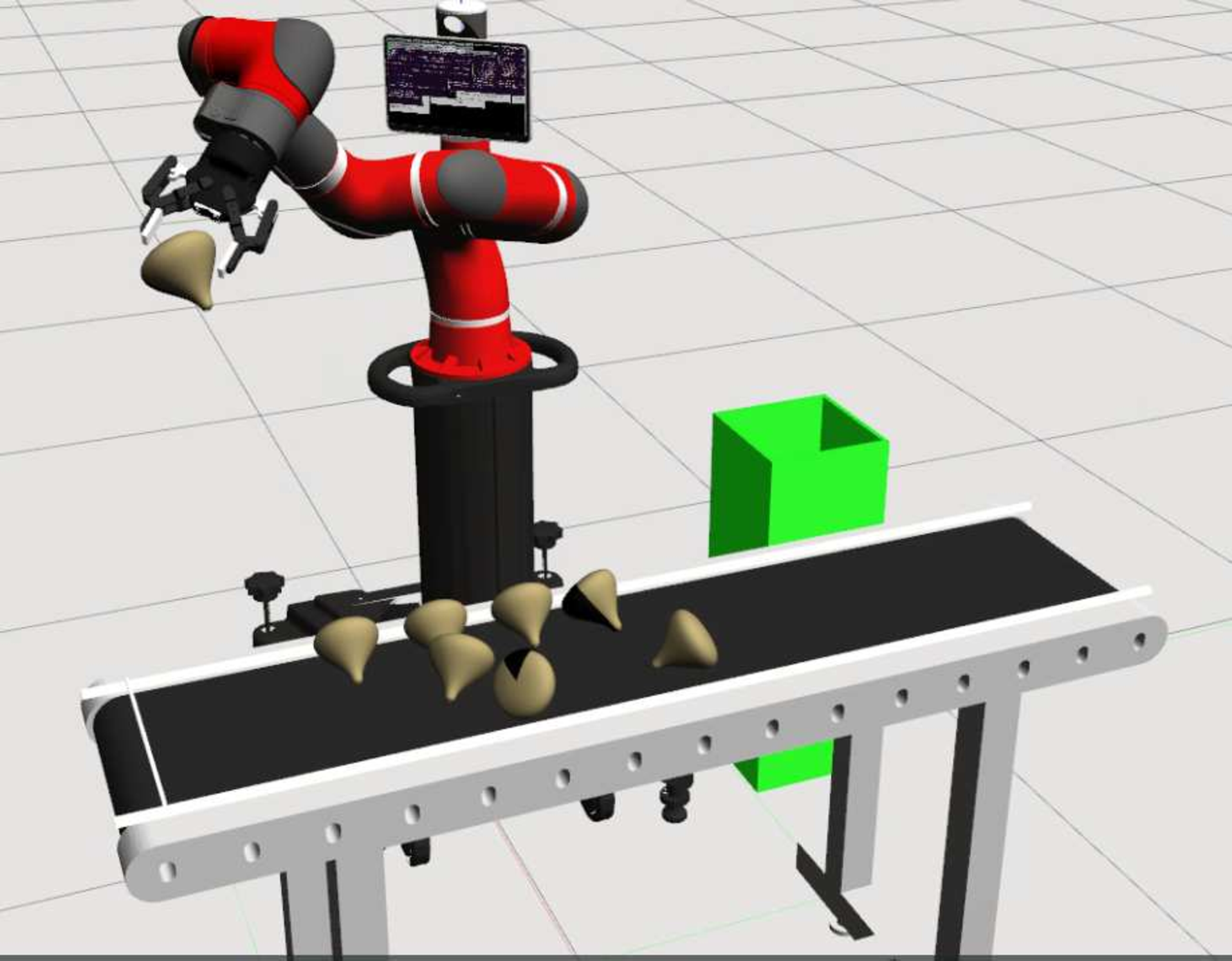}
			\includegraphics[width=0.116\textwidth,
			height=0.116\textwidth]{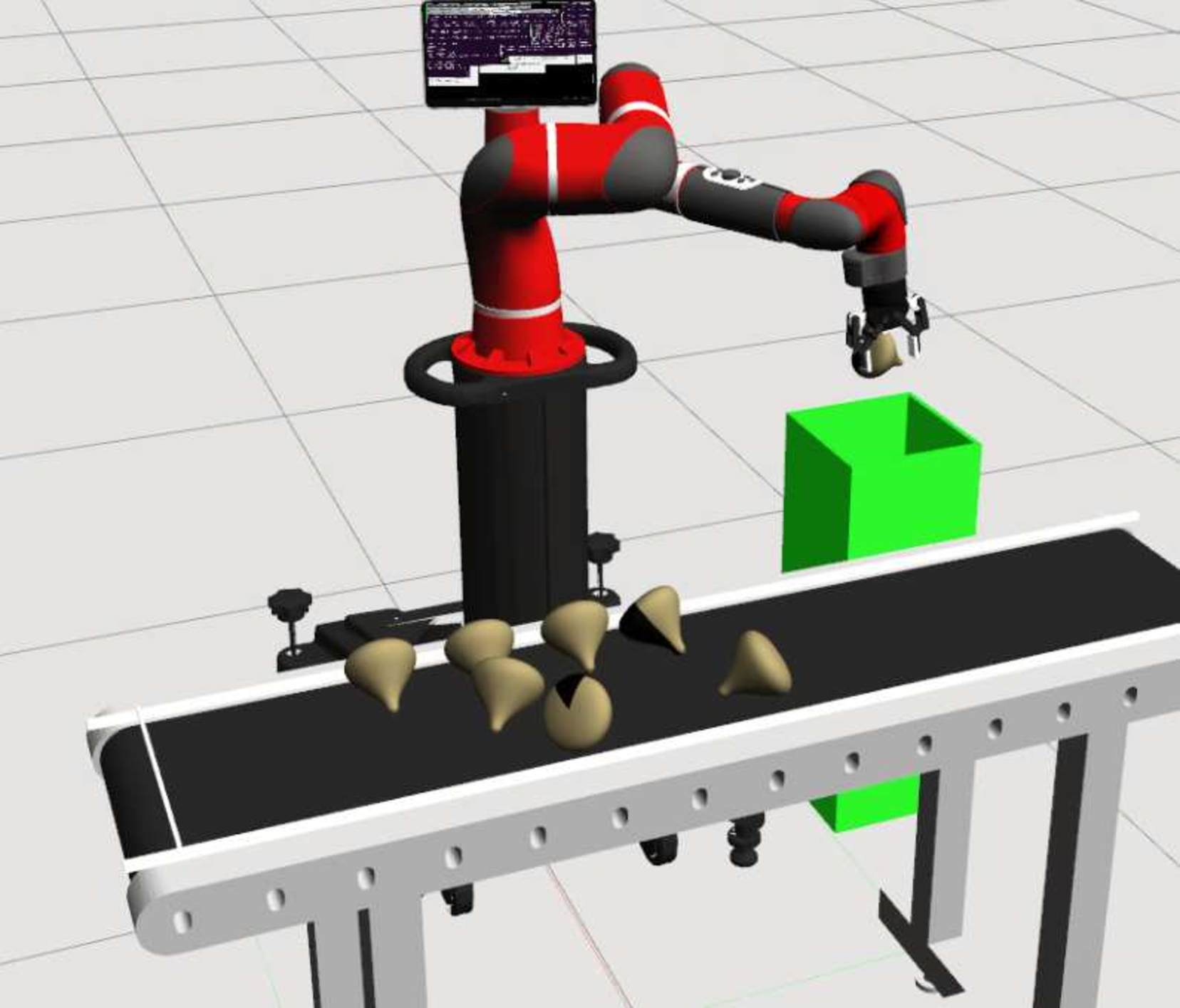}   }
			\centerline{$(a)$}
\centerline{                 \includegraphics[width=0.12\textwidth,
			height=0.12\textwidth]{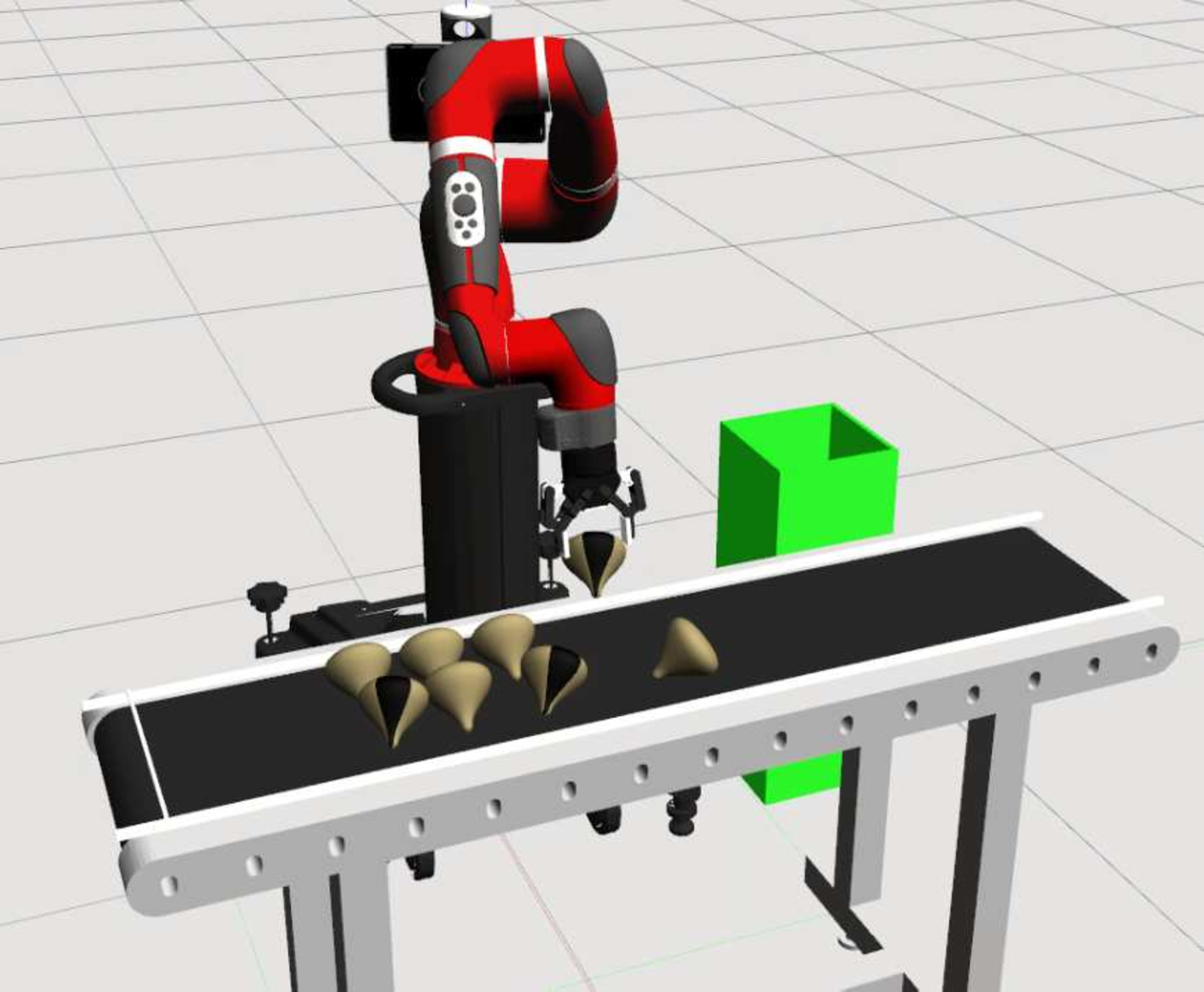}
			\includegraphics[width=0.12\textwidth,
			height=0.12\textwidth]{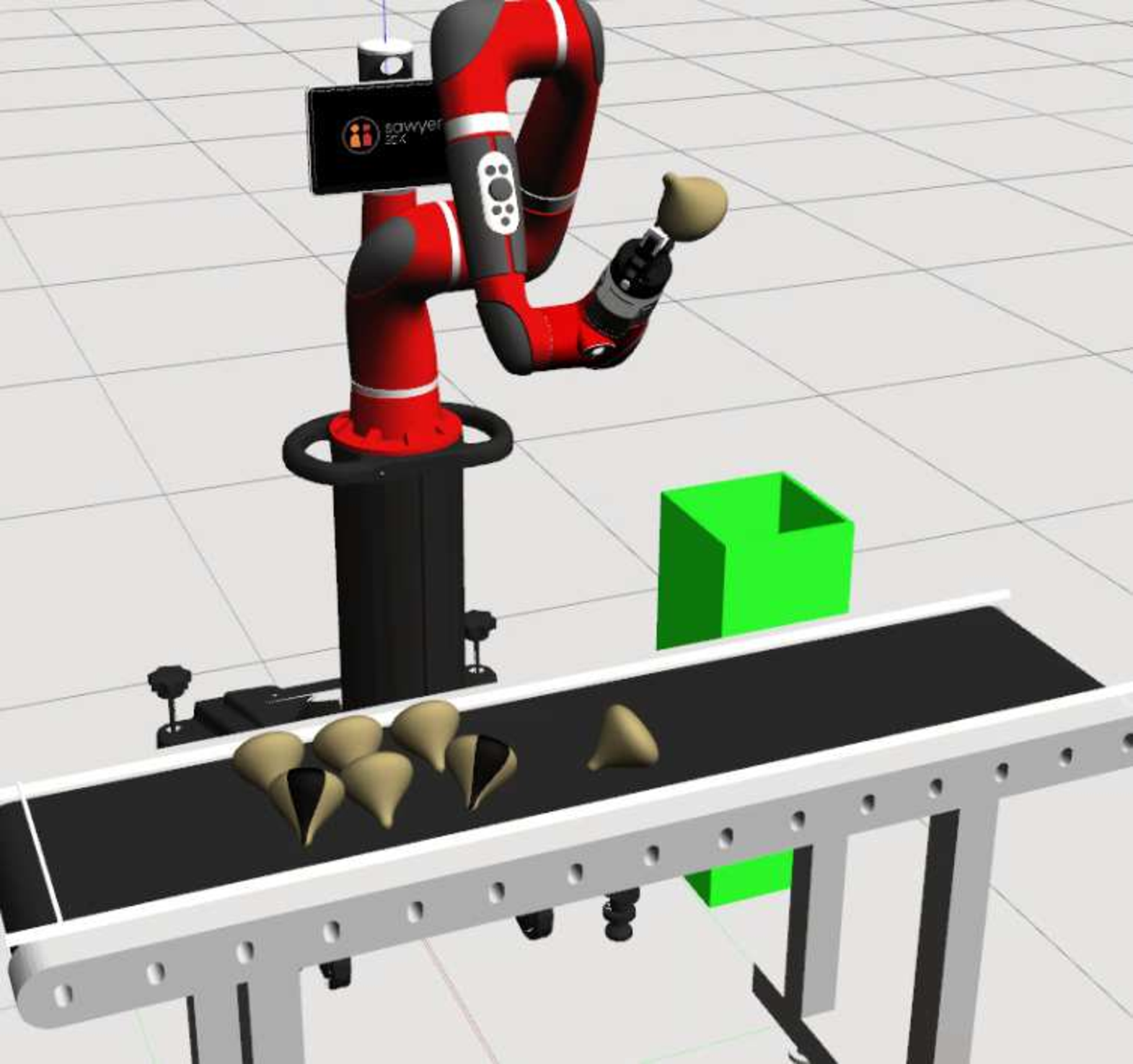}
			\includegraphics[width=0.12\textwidth,
			height=0.12\textwidth]{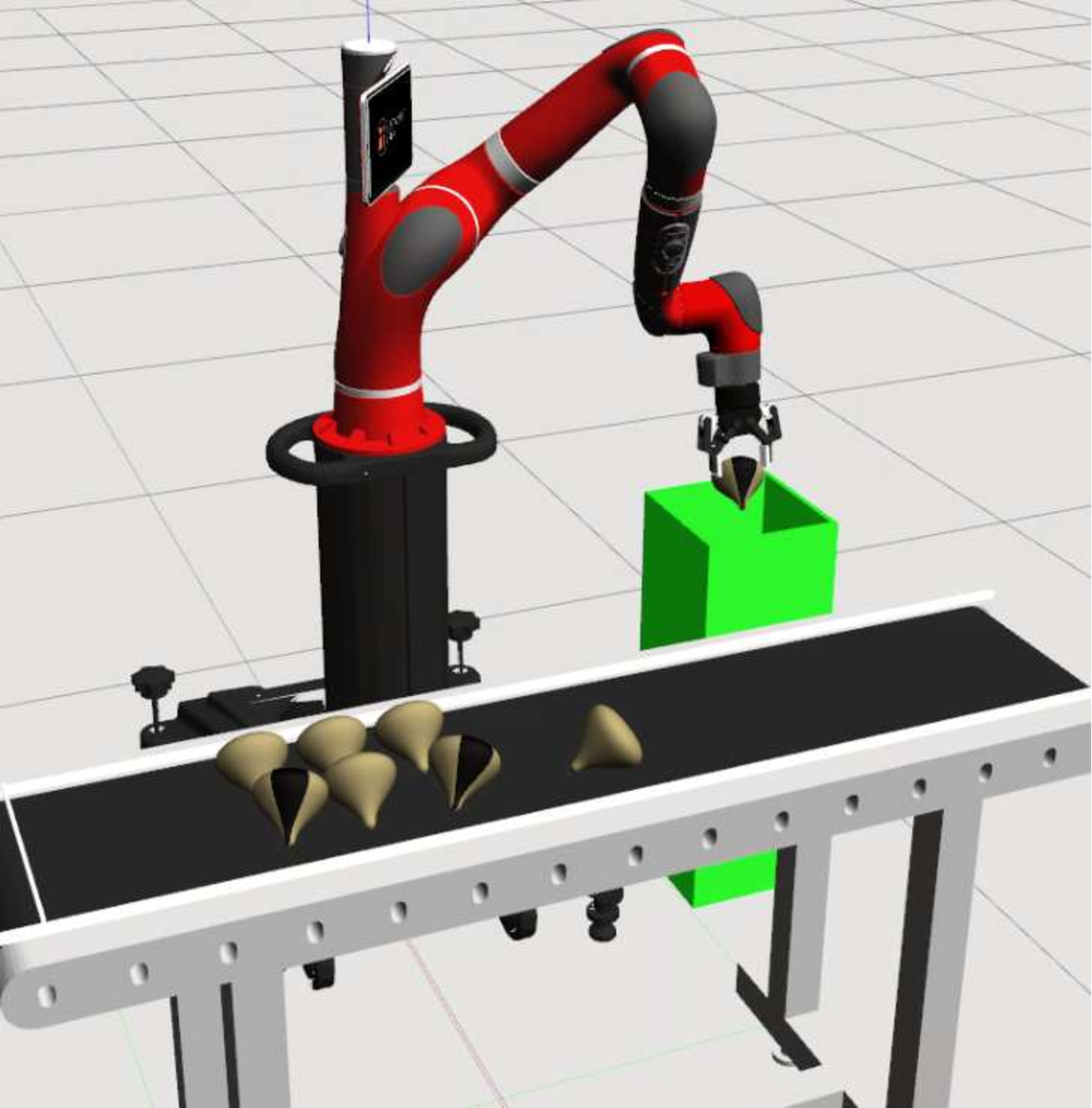}        }
		\centerline{$(b)$}
\caption{{\small A demonstration  of the  two onion  sorting behaviors  by the
		expert  in ROS  Gazebo.   $(a)$  The Sawyer  robotic  arm rolls  its
		gripper over the onions thereby exposing more of their surface area.
		Possibly blemished  onions are  then picked and  placed in  the bin.
		$(b)$ Sawyer picks  an onion, inspects it closely to  check if it is
		blemished, and places it in the bin on finding it to be blemished.}}
	\label{fig:Sawyer_sorting}	
\vspace{-0.12in}
\end{figure}

In a visit  to a real-world onion processing line  attached to a farm,
we observed  that two distinct  sorting techniques were in  common use
and would be interleaved by
the human sorters. Subsequently, we model
the expert as acting according to the output of two MDPs both of which
share  the state  and action  sets,  the transition  function and  the
reward feature functions.  They differ  in the weights assigned to the
features, which yields different  behaviors.  The specific task
is to  learn the reward  functions underlying  the two MDPs.

The  state of  a sorter is  perfectly observed  and
composed of four factors: {\em onion} and {\em gripper location}, {\em
	quality  prediction}, and  {\em  multiple  predictions}.  Here,  the
onion's  location  can be  on  the  sorting  table, picked  up,  under
inspection  (involves  taking  it  closer to  the  head),  inside  the
blemished-onion bin, or the onion has  been returned to the table post
inspection.  {\em Gripper locations} are  similar but does not include
the return back  to the table.  {\em Quality prediction}  of the onion
can  be blemished,  unblemished,  or  unknown.  Finally,  simultaneous
quality prediction for multiple onions is either available or not.

The expert's actions involve focusing attention  on a new onion on the
table at random, picking it up,  bringing the grasped onion closer and
inspecting it,  placing it in the  bin, placing it back  on the table,
roll its  gripper over  the onions,  and focus  attention on  the next
onion among those whose quality has been predicted.
The features utilized  as part of reward functions are:
\begin{itemize}[leftmargin=*,topsep=0in,itemsep=0in]
	\item {\em BlemishedOnTable}($s$,$a$) is 1  if the considered onion is
	predicted to be blemished and it is on the table, 0 otherwise;
	\item {\em GoodOnTable}($s$,$a$) is 1 if the considered onion is
	predicted to be unblemished and is on the table; 0 otherwise
	\item {\em BlemishedInBin}($s$,$a$) is 1 if the considered onion is
	predicted to be blemished and is in the bin; 0 otherwise
	\item  {\em  GoodInBin}($s$,$a$)  is  1 if  the  considered  onion  is
	predicted to be unblemished and is in the bin and 0 otherwise.
\end{itemize}
\begin{itemize}[leftmargin=*,topsep=0in,itemsep=0in]
	\item {\em MakeMultiplePredictions}($s$,$a$) is  1 if the action makes
	predictions for multiple onions simultaneously, and 0 otherwise;
	\item {\em InspectNewOnion}($s$,$a$) is 1 if the considered onion is 
	inspected for the first time and a quality prediction is made for it;
	\item {\em  AvoidNoOp}($s$,$a$) is  1 if the  action $a$  changes the
	state of the expert, 0 otherwise.
	\item {\em PickAlreadyPlaced}($s$,$a$) is 1  if the action $a$ picks
	an  onion  that  has  already  been  placed  after  inspection,  0
	otherwise. This helps avoid pick-place-pick cycles.
\end{itemize}

Two distinct vectors of real-valued weights on these feature functions
yield two distinct reward functions. The  MDP with one of these solves
to obtain a  policy that makes the expert randomly  pick an onion from
the table, inspect it  closely, and place it in the  bin if it appears
blemished  otherwise place  it back  on the  table. The  second reward
function yields  a policy that has  the expert robot roll  its gripper
over the onions,  quickly identify all onions and place few onions  in the bin if
they seem blemished. Both these  sorting techniques are illustrated in
simulation in Fig.~\ref{fig:Sawyer_sorting}.

\section{Experiments}
\label{sec:experiments}

We simulated  the domain described in  Section~\ref{sec:domain} in the
3D  simulator Gazebo  7  available as  part of  ROS  Kinetic with  the
robotic  arm {\bf  Sawyer} functioning  as a  stand-in for  the expert
grader. Sawyer is a  single robot arm with 7 degrees  of freedom and a
range of about 1.25m. 
We partially simulate a moving
conveyor belt in Gazebo by repeatedly making a collection of onions --
some of these are blemished -- appear  on the table for a fixed amount
of  time after  which the  onions  disappear.  Sawyer  is tasked  with
sorting  as many  onions as  possible from  each collection  before it
disappears.
The  locations of  onions are  internally  tracked and  made
available to the expert to facilitate  the task. We utilize the MoveIt
motion planning  framework to  plan Sawyer's various  sorting actions,
which are then reflected in Gazebo.\\

\noindent  {\bf Metrics}~~  A known  metric  for evaluating 
\irl{}'s performance    is   the    {\em    inverse    learning
	error}~\cite{Choi11:Inverse}, which  gives the loss of  value if the
learner uses the policy obtained by  solving the expert's MDP with the
learned reward function (parameterized  by $\bm{\theta}^L$) instead of
the expert's true  policy obtained by solving its MDP  with its actual
reward function 
$$ILE = \frac{1}{|\Ycal|}\sum\nolimits_{i=1}^{|\Ycal|} \|V^{\sigma^{\bm{\theta}^*_{c_i}}}(s)
-           V^{\sigma^{\bm{\theta}^L_{c_i}}}(s)\|_p.$$           Here,
$\bm{\theta}^*_{c_i}$  and  $\bm{\theta}^L_{c_i}$  are  the  true  and
learned reward function weights (component parameters) for the cluster
assigned  to  the  observed trajectory  $y_i$  and  $\sigma^{\{\cdot\}}$
denotes the  corresponding policy.  Note  that ILE averages  the value
loss    over    the    observed    trajectories.

Another pair of  metrics is used to measure the  performance of Sawyer
on  the  onion  sorting  using the  learned  reward  functions.   {\em
	Precision} is measured  as the proportion of the  number of actually
blemished onions in the bin from  the total number of onions placed in
the  bin. {\em  Recall} is  the proportion  of the  actually blemished
onions placed in  the bin from the total number  of actually blemished
onions on the  table. As careful inspection tends to  be more accurate
than  simply  rolling over  the  onions,  we  expect the  behavior  of
pick-inspect-place  to  exhibit a  higher  precision  compared to  the
alternative. On the other hand, it is slower compared to rolling and
placing, hence its recall is expected to be lower.

\begin{figure}[!ht]
	\centerline{
	\includegraphics[width=0.47\textwidth,height=0.2\textwidth]{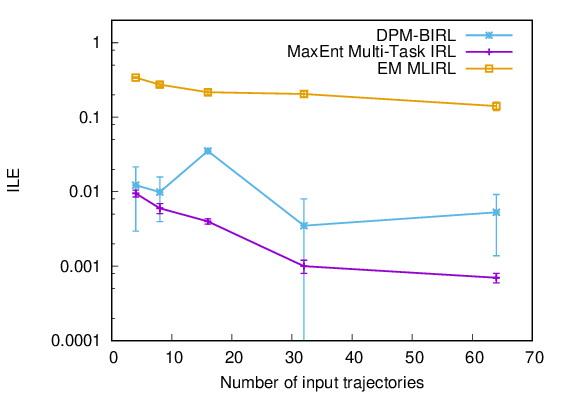}
	} 
	\caption{ {\small Average ILEs of ME-MT\irl{}, 
		DPM-B\irl{} and EM-ML\irl{} as the number  of trajectories
		increases.  Vertical bars are the standard deviations. Note these bars may have unequal height around the mean due to the
		log-scale of the y-axis.}} 
	\label{fig:results_avgEVD}
\end{figure}

\noindent {\bf Performance evaluation}~~ We  used the metric of ILE as
defined previously to measure the performance of the MaxEnt Multi-task
\irl{} (ME-MT\irl{}). Figure~\ref{fig:results_avgEVD} shows the trend of average ILE as the
number of input trajectories is increased. 
It also shows the performances for previous  multi-task  \irl{} techniques: DPM-B\irl{}~\cite{Choi12:Nonparametric} and EM-ML\irl{}~\cite{Babes-Vroman11:Apprenticeship}. Each  data point  is the average of 5 runs. In each  run, the gradient ascents and descents are allowed  to  stabilize thereby  yielding  stable  feature weights  and
cluster assignments. Note  the expected monotonic decrease in ILE  exhibited  by  ME-MT\irl{}  as the  number  of demonstration trajectories increases. Unlike the two DP based methods, EM-ML\irl{} performed significantly worse throughout converging to incorrect reward functions that were likely local optima. 
The MaxEnt method (ME-MT\irl{}) exhibits ILE that  is  consistently  lower  than  that of the Bayesian method (DPM-B\irl{}),  which  is indicative of  better learning  performance. Notice that 64 trajectories appear to be sufficient to get a very low ILE. Finally, the method learned that there were two clusters in most runs though a few yielded just one cluster.

\begin{table}[!ht]
	\begin{small}
		\begin{tabular}{p{0.06\textwidth}|p{0.14\textwidth}|p{0.11\textwidth}|p{0.07\textwidth}}
			\hline 
			& {{\bf Method}} & {{\bf (TP,FP,FN,TN)}} & {{\bf P\%,R\%}}\\
			\hline
			Expert & Pick-inspect-place & (4,0,8,12) & 100, 33 \\  
			& Roll-pick-place & (8,4,4,8) & 66, 66 \\ \hline \hline
			Learned & Pick-inspect-place & (3,0,9,12) & 100, 25 \\  
			(ME-MTIRL) & Roll-pick-place & (6,4,6,8) & 60, 50 \\ \hline
			Learned & Pick-inspect-place & (2,0,10,12) & 100, 16.7 \\  
			(DPM-BIRL) & Roll-pick-place & (5,5,7,7) & 50, 41.7 \\ \hline 
			Learned & Pick-inspect-place & (3,1,9,11) & 75, 25 \\  
			(EM-MLIRL) & Roll-pick-place & (5,4,7,8) & 55.6, 41.7 \\ \hline 
		\end{tabular}
	\end{small}
	\caption{{\small Column labels  TP denotes true positive  (\# blemished onions
		in  bin), FP  denotes false  positive (\#  good onions  in bin),  TN
		denotes true negatives  (\# good onions remaining  on conveyor), and
		FN  denotes  false  negatives  (\#  blemished  onions  remaining  on
		conveyor). P and R denote  precision and recall in \%, respectively.}}
	\label{tab:precision_recall}
\end{table}

However, does the improvement in learning translate to improved performance in the sorting task? In Table~\ref{tab:precision_recall}, we show the average precision and recall  of the  expert engaged  in using the two sorting   techniques and  the analogous metrics for the learned  behaviors using all three \irl{} approaches. We used the average  feature weights  learned across  5 trials. The performance of the behaviors learned by ME-MT\irl{} is closer to that of the expert's than those learned by the two baseline methods. Notice that  the learned pick-inspect-place  behavior shows  high precision  but leaves many onions on  the table leading to worse recall. It sometimes gets trapped in a cycle where Sawyer repeatedly picks and places the same  onion on  the table. This  is because a  very low  weight is learned for  the {\em PickAlreadyPlaced($s$,$a$)} feature  function, a high weight  for this feature would  have avoided this cycle. On the other hand, the roll-pick-place behavior is learned satisfactorily and
exhibits precision and recall close to those of the true behavior.

\section{Related Work}
\label{sec:related}

An early work  on  multi-task   preference  learning~\cite{Birlutiu09:Multi-task} represents the relationships among various preferences  using a    hierarchical   Bayesian model.   Dimitrakakis    and   Rothkopf~[\citeyear{Dimitrakakis12:Bayesian}]
generalized this  problem of  learning from  multiple teachers  to the
dynamic  setting   of  IRL,  giving  the   first  theoretically  sound
formalization of  multi-task IRL. They  modeled the reward  and policy
functions as being drawn from a  common prior, and placed a hyperprior
over the common  prior. Both the prior and the  hyperprior are updated
using  the  observed  trajectories  as evidence,  with  the  posterior
samples capturing the distribution  over reward functions that explain
the  observed  trajectories.  The inference  problem  is  intractable,
however, even for toy-sized MDPs.

The above setting  of multi-task IRL was also restricted  in the sense
that this was  the ``labeled'' variant; although  the reward functions
generating  the trajectories  were unknown,  it was  assumed that  the
membership of  the observed trajectories  to the set generated  from a
common     reward    function     was    known.     Babes-Vroman    et
al.~[\citeyear{Babes-Vroman11:Apprenticeship}]    first     addressed    the
``unlabeled'' variant of multi-task IRL, where the pairing between the
unknown  reward  and the  trajectories  was  also unknown.  They  used
expectation  maximization  to  cluster  trajectories,  and  learned  a
maximum likelihood reward function for each cluster. By contrast, Choi
and  Kim~[\citeyear{Choi12:Nonparametric}] took  the  Bayesian IRL  approach
using  a Dirichlet  process mixture  model, performing  non-parametric
clustering  of  trajectories,  also  allowing  a  variable  number  of
clusters  for  the  ``unlabelled'' variant.  Both  approaches  treated
multi-task IRL as multiple single-task IRL problems, which is also the
viewpoint      taken      in       our      work.      Gleave      and
Habryka~[\citeyear{Gleave18:Multi-task}]    take   a    markedly   different
viewpoint: they  assume that the  reward functions for most  tasks lie
close to the mean across all tasks. They propose a regularized version
of MaxEnt IRL that exploits  this assumed similarity among the unknown
reward functions, thus transferring information across tasks. Although
it is also an application of MaxEnt IRL to the multi-task setting, our
approach is a principled integration  of the Dirichlet process mixture
model into MaxEnt IRL.

The  problem of  learning from  multi-expert demonstrations  have also
been  studied  from  a  non-IRL  perspective,  specifically in imitation learning,   such   as the  generative  adversarial   imitation   learning (GAIL)~\cite{Ho16:Generative}.       Recent        extensions       of GAIL~\cite{Hausman17:Multi-modal,Li17:InfoGAIL}  augment  trajectories with a latent intention variable specifying  the task, and then maximize the mutual information between observed trajectories and the intention variable  to disentangle  the trajectories generated from  different tasks. On the other hand,  GAIL's performance has been difficult to replicate and MaxEnt\irl{} has been shown to perform better than GAIL on single tasks~\cite{Arora19:LMEI2RL}.


\section{Concluding Remarks}
\label{sec:conclusion}

Humans may exhibit multiple distinct ways of solving a problem each of which optimizes a different reward function while still sharing the  features. For \irl{} to remain useful in this context, it must be   generalized to not only learn how many reward functions are being 
utilized in the demonstration, but also to learn the parameters of these reward functions. We presented a new multi-task \irl{} method that combined maximum entropy \irl{} -- a key \irl{} technique -- with elements of the Dirichlet process Bayesian mixture model. While keeping the number of learned behaviors (minimum entropy clustering) minimal necessary to explain observations, it allows to leverage the advantages of maximum entropy \irl{} and facilitates solving this generalization directly as a unified optimization problem.  On a real-world inspired domain, we showed that it improves on the previous  multi-task \irl methods. The behaviors induced by the learned reward functions imitated the observed ones for the most part.

Having established the value of combining MaxEnt with multi-task \irl{} in this paper, our next step is to explore how well this method scales to problems with more actions and tasks. An avenue of ongoing work aims to derive sample complexity bounds by relating the optimization to a maximum likelihood problem. Such bounds could inform the number of trajectories needed for a given level of learning performance. 

\newpage

\bibliographystyle{named}
\bibliography{ijcai20}

\end{document}